# Speech Recognition Oriented Vowel Classification Using Temporal Radial Basis Functions

Dr. Mustapha GUEZOURI, Larbi MESBAHI and Abdelkader BENYETTOU
Signal Image Parole Laboratory, Department of Computer Sciences, Faculty of Sciences,
University of Science and Technology (USTO),
P.O. Box 1505 – EL-M'NAOUER, Oran, ALGERIA.

**Abstract-** The recent resurgence of interest in spatio-temporal neural network as speech recognition tool motivates the present investigation. In this paper an approach was developed based on temporal radial basis function "TRBF" looking to many advantages: few parameters, speed convergence and time invariance. This application aims to identify vowels taken from natural speech samples from the Timit corpus of American speech. We report a recognition accuracy of 98.06% in training and 90.13 in test on a subset of 6 vowel phonemes, with the possibility to expend the vowel sets in future.
**Keywords:** Speech Recognition, TRBF, TDNN, Vowel phonemes, Timit corpus.

## 1 INTRODUCTION

In recent years, the advent of new learning procedure and the availability of high performance computers have given rise to a renewed interest in connectionist models of intelligence [1].

More precisely, time takes an advanced place in the recent intelligent systems, where the notion of time delay becomes a useful characteristic in the dynamic representation of shapes, and this delay can bind the past, the present and the future for better decision.

However, the speech recognition systems employed in personal speech recognition software and even in more advanced systems tend to use Hidden Markov Models either in their place or in conjunction with them [2]. However, HMMs are not equally biologically plausible and have significant shortcomings [2]. Many contributions give a great interest in artificial neural network field because it has demonstrated that connectionist architectures are capable of capturing some critical aspects of the dynamic nature of speech, can achieve superior recognition performance for difficult but small phonemic discrimination tasks such as discrimination of the voiced consonants /B/, /D/ and /G/ [3]. But the main problem of classical methods was the hard time processing and the adjustment of parameters that become a laborious stain for the large vocabulary in continuous speech. In opposite, the RBF networks don't require a special adjustment and the training time becomes shorter with regard to the TDNN. But the problem is the shift invariance in time [4].

The goal to combine the approach of the RBF with the shift invariance features of TDNN can be get a new robust model, this is named temporal radial basis function "TRBF" [1], and to be more efficient, we have adapted the time delay for these networks in object to be more dynamic according to vowel phoneme behaviour has study. The approach designed by the authors are trained and tested on data pre-processed using mel frequency cepstral coefficient "MFCC" feature analysis. Thus the role of the neural network here is to provide a mapping from 13-dimensional MFCC in space of vowel phonemes. The novelty consists to create according to incremental learning a hidden neuron block instead of only one neuron in the hidden layer each iteration; in addition it adjusts the size of this block correspondingly to size of input time delay which characterizes the specialized window. Commonly we have kept the advantages of time delay neural networks "TDNN" and its capacities to detect the acoustic features and their independent temporal report during learning, except we have discarded the shared weight notions.

## 2 THEORETICAL FOUNDATION

The vowel classification procedure described in this paper is based on three fundamental concepts: the mel-frequency cepstral coefficients (MFCCs), self organizing map quantization, and temporal radial basis functions neural networks "TRBF".

### 2.1 Mel Frequency Cepstral Coefficients (MFCCs)

A classic graphical description of an audio signal, especially one generated by speech, is the so-called "voiceprint". An audio signal is a one-dimensional stream of data, i(t): for each point in time, it describes an acoustic pressure. We wish to describe the "instantaneous" frequency spectrum at some time T (see figure1); this is a vertical slice of the spectrogram. Since the Fourier transform only works on a sequence of data points, we sample a short segment of the signal about time T by multiplying it by a windowing function [5]. One common windowing function is the Hamming window, given by the following equation (1), the normalization factor at the left of the Hamming window equation is chosen so that the mean square of the function is 1[5].

$$W(t) = \frac{46}{\sqrt{1691/2}} * (\frac{25}{46} - \frac{21}{46} * \cos(2\pi t)) \quad (1)$$

Performing a Fourier transform on the windowed signal gives us a snapshot of the frequencies in the signal at time T. To create the spectrogram, we simply move T along the length of the signal in small steps, windowing the signal and generating a spectrum at each step. The spectrogram is a collection of all these spectra arranged side-by-side. Dealing directly with the massive data set that the spectrogram represents is obviously not a practical way to algorithmically analyze a speech sample [5].

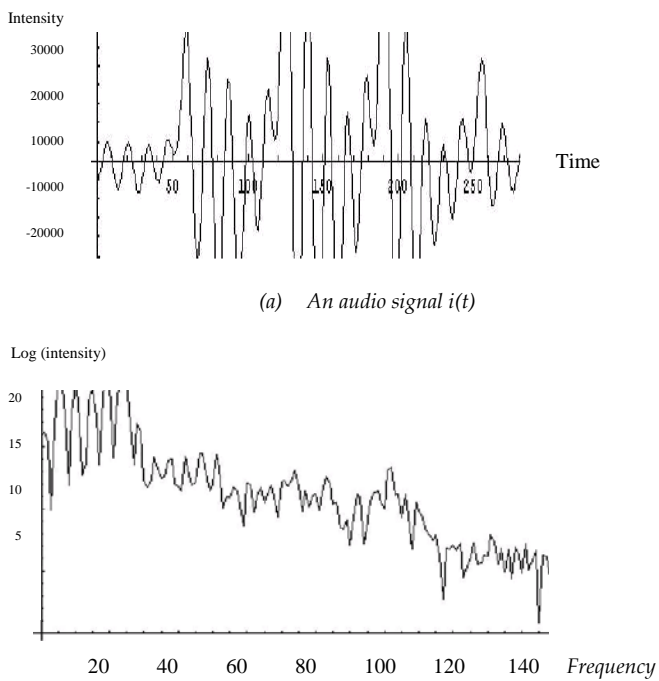

(a) An audio signal i(t)

(b) The spectrum of the signal i(t)
Fig.1. An audio signal (a) and its spectrum, log-scaled (b)

We would instead prefer a compact summary, in the form of a relatively low-dimensional feature vector, of the perceptually meaningful characteristics of the sound at each point in time. Experimentally established feature vectors include linear prediction coefficients and mel-frequency cepstral coefficients [5]; we choose the latter of these.
The mel frequency scale grew out of the early observation that humans can differentiate low-frequency sounds better than high-frequency ones; the gaps between tones that listeners judge to be equally spaced grow larger as frequency increases. Thus the mel scale, in which the distance between tones is approximately linear with their perceptual distance, is a logarithmic transform of the ordinary frequency scale. MFCCs are thus the result of a Fourier transform of the log warped frequency spectrum, and can be plotted over time in a low-resolution plot similar to a spectrogram.

## 2.2 Self organizing map quantization

Self-organizing in networks is one of the most fascinating topics in the neural network field. Such networks can learn to detect regularities and correlations in their input and adapt their future responses to that input accordingly [6]. The neurons of competitive networks learn to recognize groups of similar input vectors. Self-organizing maps learn to recognize groups of similar input vectors in such a way that neurons physically near each other in the neuron layer respond to similar input vectors (see figure 2).

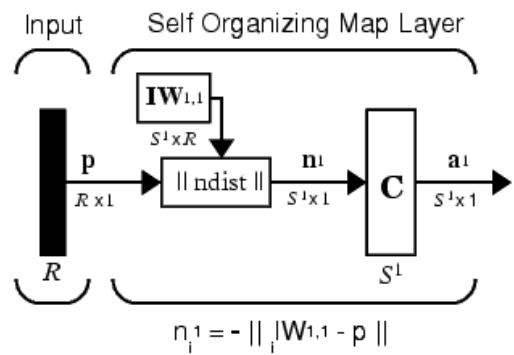

$$n_i^1 = - \| {}_iIW_{1,1} - p \|$$

R : *number of element in input vector*
S1: *number of competitive neurons.*
n1: *the distance vector,* a1: *the winner*
IW: *is the weight matrix*

Fig. 2. Self Organizing Map for quantization

A self-organizing map learns to categorize input vectors. It also learns the distribution of input vectors. Feature maps allocate more neurons to recognize parts of the input space where many input vectors occur and allocate fewer neurons to parts of the input space where few input vectors occur. Self-organizing maps also learn the topology of their input vectors.
The aim is to reduce huge and massive data (large vocabulary corpus) to a similar example basis which the cardinality represents a compromise between a minimum of loss information and minimum of time processing [1].

## 2.3 Temporal radial basis function approach

The classic RBF can be formed to accomplish tasks of the pattern recognition with no linear and complex contours

[7], they are limited to treat some static models, rather than to treat shapes that are in temporal nature. The training of such network requires some techniques to adapt the hidden neurons with passages of input windows so that it is an adequate integration of information according to the temporal interval.

The Temporal RBF [1], ATDNN [8],..etc, all these approaches are proposed to defeat the static limitation. Networks with this capacity can play an important role in applications domain, possessing some dynamic properties in pattern recognition for example: the no- stationary signals and the dynamic shapes [4], [9], [6].

Also to take part of classic RBF advantages in approximation and recognition, the objective come closer toward a behaviour wanted by a collection of kernels [4]. In this sense the architecture is able to adapt to the time-scale of the input tokens and there is no need to specify the architectural constraints [10] (see Fig. 1), we note:

1) The width σ of the Gaussian shaped input window determines how much temporal context each connection transmits.
2) The number of connections is allowed to increase during learning. Thus, the training algorithm determines how many independently trainable weights are needed to capture the temporal dynamics within the required temporal context.
3) The adaptive time delays were already realized what temporal context unit receives (that means how far each unit looks back into the past).

With similarity to TDNN network, TRBF must be able to:
- Represent temporal relationships between acoustic events, while at the same time.
- Provide for invariance under translation in time [10].

Also our approach can approximate any spatiotemporal function according to the desired accuracy by using the Stone-Weirstrass theorem [8], under some restrictions:
- If the kernel $\phi$ is a bounded and monotone increasing differentiable functions.
- At least one hidden layer of N hidden block units.
- Having d-times delays elements in each input hidden connections pairs.

**2.3.1 Network Architecture for Phoneme Recognition**

To allow a RBF network detecting features in time, it is necessary to not only present inputs to one point data, but in many passages. In this case, we take a structure of a classic RBF with some modification; instead to take a hidden neurone we took a block of hidden neurones.

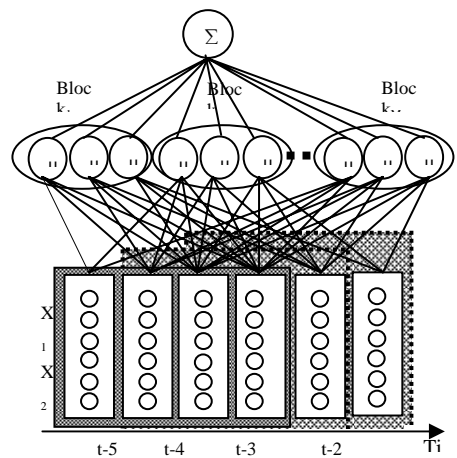

*Fig. 3. Parameters of this TRBF network are:*
*n: number of features equal to dim(X)=6, Nde : Time delay equal to 4, Nfe: Size of the input window, here Nfe= 6, Nnc : number of hidden neurons by block, here Nnc= 3 , Nbc: represents the number of centres in hidden bloc, $\mu$: represents a centre , dim($\mu$)= Nde x n.*

We notice in this network that the input vector have a Shifted window with $\Delta t$ length, what burdens the time delay count between centres and inputs.

In this type of architecture, the goal is to get an elevated precision, in addition we can adapt for every shape the network while changing the number of block in the hidden layer and this is in minimizing the size of the block of input time delay. Therefore the only parameters to adjust are: the size of input windows (Nfe) and the time delays (Nde) and the number of blocks in the hidden layer.

We notice that the taken time delay number for the input vector depends on the application, generally in word recognition; the size of the observation window vectors is in the order of 10 to 15, therefore the number of time delay is acceptable between 5 and 10. Concerning the number of blocks of the restrained RBF in the hidden layer is fixed either according to the wanted training rate or following error thresholds (Akaike criterions) [1].

We make remark on speed of training that characterizes our type of network in comparison to other architectures like back-propagation or feed forward approaches. In opposite to TDNN here, the weights are not shared, this feature improve the time convergence in learning phase.

**2.3.2 Dynamic OLS Algorithm**

The orthogonal least square "OLS" with its classic version can not adapt our TRBF network, therefore the original idea of the Dynamic OLS method resides at each iteration, the creation of a hidden centre block and not only one centre, as the size of each block is expressed like suit :

If we consider that the input vector is composed of n characteristic on a temporal input window with Nfe length and if we take the value of input time delay equal to Nde, such as: Nfe ≥ Nde. Then the number of neurons composing every block is equal to: Nfe-Nde+1.

The size of every hidden centre is equal to: n x Nde, (see Fig. 1, for more clarity).

Also this algorithm permits to make an incremental training [7] in following steps:
1) First it makes the linear separation between the input layer and the hidden layer; it creates the hidden neurons automatically while applying the Gram-Schmidt orthogonalisation, which permits to eliminate redundancies of information. Other methods consist in using the genetic algorithms to minimize the number of hidden neurons with a good generalization [1].
2) Secondly to make the training between the hidden layer and the output layer, using the least square method while calculating synaptic weights.
3) With this representation, we can see the following steps:

$$d = (P * \theta + E) \quad (2)$$

The orthogonalisation of the columns Pi can be gotten by the decomposition of the P matrix in two matrices W and A as:

$$P = W A \quad (3)$$

Where W: of size Nex M, is the orthogonal image of the P matrix. A: of size M x M, is a superior triangular matrix containing orthogonal coefficients. The A matrix is defined by equa. 6.

The space begotten by the vectors Pi is the same space begotten by the vector W, then

$$d = (W * A * \theta + E) \quad (4)$$

Where G=A. θ is the search solution.

$$[err]_i = G_i^2 \frac{W_i^T W_i}{d^T d} \quad (5)$$

By iteration we calculate the elements of A and W

$$\alpha_{j,k}^i = \frac{W_j^t * P_i}{W_j^t * W_j} \quad (6)$$

$$W_k^i = P_i - \sum_{j=1}^{j=k-1} \alpha_{j,k}^i * W_j \quad (7)$$

The criterion of iteration stop here is not based merely on the Akaike criterion but by adding the (Eq. 9) to formulate the hidden block size in hidden layer.

$$1 - \sum_{i=1}^{i=M} err_i \leq \varepsilon \quad (8)$$

Modulo M on Nnc=0, where M≠0 (9)

In end of iterations, we calculate the synaptic weights according to the system:

$$G = A * \theta \quad (10)$$

## 3 EXPERIMENTAL PARTY

The developed approach has been achieved on a subset of the TIMIT data base [11] organized in 6 vowels, 6 fricatives and 6 plosives. For our survey we reduced the space of study for the case of vowels (see Table. 1).

### 3.1 Data basis and training

Signals have been sampled to 16 khz with an analysis cepstral under the Mel scale, takes all 20ms in Hamming windows of 25ms giving each 12 MFCC coefficients and the corresponding residual energy (13th parameter).

|        | Train *set* | Test set |
|--------|-------------|----------|
| /ah/   | 2200        | 879      |
| /aw/   | 700         | 216      |
| /ax/   | 3352        | 1323     |
| /ax-h/ | 281         | 95       |
| /uh/   | 502         | 221      |
| /uw/   | 536         | 170      |

TABLE 1
SUBSET OF TIMIT BASE
CONTAINING THE /AH/, /AW/, /AX/, /AX-H/, /UH/ AND /UW/

This work was achieved on a PIV microcomputer 2.8 GHz with 256 Mo of RAM, developed by the C++ builder and Matlab 7.0 languages. Concerning the training data basis, it has been quantified with a SOM-Kohonen with size 16 x 16 [11], generating a basis of 256 phoneme examples, therefore the size of training basis is about 1500 examples (250 example for each vowel), every example is normalized on an input length window equal to 5. The size of the test basis is 750 examples with mean of 125 examples by phoneme.

The parameters of our TRBF are : Ne=1500, Nfe=5 , Nde can take the values between 1 and 5, Nnc=Nfe-Nde+1, Nbc varies according to phoneme basis training and the precision, the used kernel is gaussian with receiving field equal to 1, the threshold error is fixed to 0.1 and finally the data are preprocessed by center-reduce method.

In general after 60 iterations we can construct the neural network separately for each phoneme with a number of hidden neuron blocks equivalently to number of iterations.

### 3.2 Results and comments

#### 3.2.1 Effect of Time Delay Changes

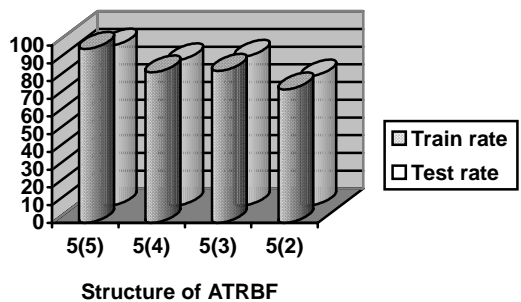

*Fig. 4. Performance according to size of windows x(y), x: Nfe, y: Nde.*

The global rate accuracy is about 98% in learning and 90.13% in test; the Fig. 4 shows us the influence of the time delay changement on performance of the network. In this case we remark that performances decrease from 5(5) to 5(2), may be this is due to insufficiency of number of input window in sweeping the speech signal spaces. In this case we can not judge the effect of variability of structure on results. One solution is proposed, it consists to try for large input window in range of [10, 15] instead of only 5.

### 3.2.2 Confusion matrix

First we have 250 examples for each phoneme in learning and 125 examples for test. When we observe the confusion matrices for both training and test phases, we conclude that the principal confusion is caused by the 5th vowel phoneme "/uh/", for example in table 3, it take 8 examples from /ah/, 2 from /aw/, 9 from /ax/, 1 from /ax-h/, and 18 from /uw/.

TABLE 2. CONFUSION MATRIX FOR LEARNING EXAMPLES

|  | /ah/ | /aw/ | /ax/ | /ax-h/ | /uh/ | /uw/ |
|---|---|---|---|---|---|---|
| /ah/ | 247 | 1 | 2 | 0 | 0 | 0 |
| /aw/ | 3 | 247 | 0 | 0 | 0 | 0 |
| /ax/ | 2 | 0 | 239 | 4 | 5 | 0 |
| /ax-h/ | 0 | 0 | 0 | 249 | 1 | 0 |
| /uh/ | 0 | 0 | 3 | 0 | 245 | 2 |
| /uw/ | 0 | 2 | 1 | 0 | 3 | 244 |

TABLE 3. CONFUSION MATRIX FOR TEST EXAMPLES

|  | /ah/ | /aw/ | /ax/ | /ax-h/ | /uh/ | /uw/ |
|---|---|---|---|---|---|---|
| /ah/ | 116 | 1 | 0 | 0 | 8 | 0 |
| /aw/ | 15 | 107 | 1 | 0 | 2 | 0 |
| /ax/ | 2 | 2 | 112 | 0 | 9 | 0 |
| /ax-h/ | 0 | 3 | 0 | 121 | 1 | 0 |
| /uh/ | 6 | 0 | 0 | 0 | 115 | 4 |
| /uw/ | 0 | 2 | 0 | 0 | 18 | 105 |

In object to improve and resolve this confusion we propose to introduce the technique of penalisation, or to change the size of input window in neural network structure.

### 3.2.3 Comparison with HMM, TDNN (different variants)

On the following Fig. 5, we are going to extricate the following comparisons:

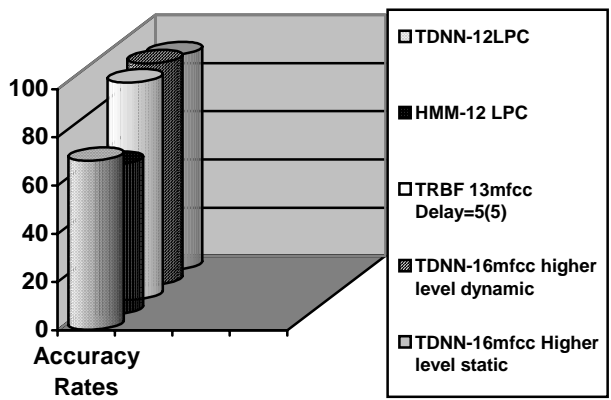

*Fig. 5. Comparison between the different approaches in Accuracy Rates*

From the Fig. 5 we remark that the TDNN with 12 LPC and HMM with 12 LPC in reference [11] gave results of 70% and 63% consecutively in test accuracy rate, this due to LPCs are less representative than MFCCs. When passing to the results of Waibel & Hirai [12], we remark that using 10 phonemes including vowels from the ATR database, and using TDNN with 16 MFCCs with two variants: static and dynamic they have obtained 89% and 92% consecutively in accuracy rates. This last takes advantage over our approach, we think that the number of input delay play a main role in this case, adds to that the number of 16 MFCCs also can be favourable for best accuracy. In general the model proposed in this paper gave good results, it comes back has hybridisation between advantages of the TDNN approach and RBF networks [4], [1].

### 3.2.4 The Noise Effect

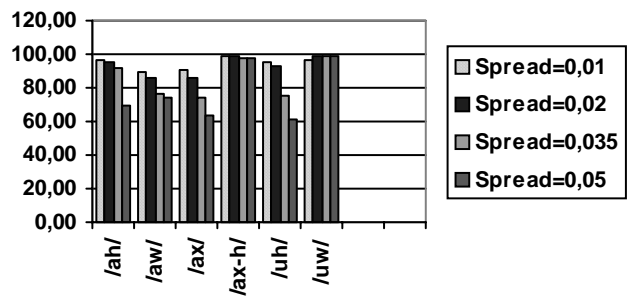

*Fig. 6. Accuracy Rate characterising the tolerance of the approach proposed towards a gaussian noise of different spread $\sigma$.*

While noticing the Fig. 6, we can deduct that the TRBF among so many other methods [13], can adjust well with a word dived in a noised middle, by adding to the test data basis a Gaussian noise of spread belongs to 0.01 until 0.035. We see that the classifier resists good until a threshold spread $\sigma$ equal to 0.035 where the test rate overcome 70% but from $\sigma$ passing the value of 0.05 the rate degrades, it can even notice itself at the human being, we doesn't sometimes manage to distinguish passages for some words if we are in very noisy environment.

## 4 CONCLUSION

In this article we have presented a new approach based on temporal radial basis function, applied to speech recognition. The main advantage with regard to other neural architectures, it is well the won time in training; in addition we have few parameters to adjust and the block size is determined automatically according to the input time delay.

The TRBF combines advantages of the TDNN that are translation invariant, where the features learned by the network are insensitive to shift in time. Examples demonstrate that the network was indeed able to learn acoustic-phonetic features, such as formant movements and segmentation, and use them effectively as internal abstractions of speech. It has shown that our TRBF has a good rate in both training and test. We suggest in future applications the introduction of genetic algorithms in computing the delay which differs from phoneme to another.

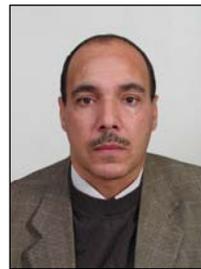

M.GUEZOURI was born in Algeria in 1962. He received the engineer degree from the national Algerian institute of Telecommunications (ITOran) in 1988. He received the Master and PhD degrees in electrical engineering in 1990 and 2007, respectively from the Science and Technology University in Oran.
Since 1990, he has been a Research Assistant at Signal and processing Laboratory, University in Oran.
He is currently an Assistant Professor of electrical and computer engineering at the University in Oran, Algeria. His current research interests include Signal processing, neural networks and Ad hoc networks.